\documentclass{article}
\usepackage{ijcai26}

\usepackage{times}
\usepackage{soul}
\usepackage{url}
\usepackage[hidelinks]{hyperref}
\usepackage[utf8]{inputenc}
\usepackage[small]{caption}
\usepackage{graphicx}
\usepackage{subcaption}
\usepackage{amsmath}
\usepackage{amsthm}
\usepackage{booktabs}
\usepackage{algorithm}
\usepackage{algorithmic}
\usepackage[switch]{lineno}

\title{Clinically Aware Synthetic Image Generation for Concept Coverage in Chest X-ray Models}

\author{
Amy Rafferty$^1$
\and
Rishi Ramaesh$^2$\and
Ajitha Rajan$^1$\\
\affiliations
$^1$University of Edinburgh, UK\\
$^2$NHS Lothian, UK\\
\emails
s1817812@ed.ac.uk
}

\begin{document}

\maketitle

\begin{abstract}
Deep learning models for chest X-ray diagnosis are constrained by limited coverage of clinically meaningful concept combinations in publicly available training datasets. While synthetic image generation has been explored to increase data diversity, existing methods rarely enforce clinical or anatomical constraints, limiting utility for improving model reliability. We propose CARPA, a clinically aware and anatomically grounded framework for synthetic chest X-ray generation that applies targeted perturbations to clinical concept vectors while preserving anatomical structure. By producing anatomically faithful synthetic images with controlled concept insertions and deletions, CARPA expands clinically relevant concept coverage. We evaluate CARPA across seven backbone architectures by fine-tuning models on synthetic subsets and testing on a held-out MIMIC-CXR benchmark. Compared to prior concept perturbation approaches, fine-tuning on CARPA-generated images consistently improves precision–recall performance, reduces predictive uncertainty, and improves model calibration. Structural and semantic analyses demonstrate high anatomical fidelity, strong concept alignment, and low semantic uncertainty. Evaluation by two expert radiologists further confirms realism and clinical agreement. Together, these results show that anatomically grounded concept perturbations enable more effective use of synthetic data, improving both performance and reliability of chest X-ray classification models and supporting safer clinical deployment.
\end{abstract}

\section{Introduction}

Deep learning models for chest X-ray (CXR) interpretation have achieved strong performance on standard diagnostic benchmarks, driven by the availability of large-scale public datasets such as MIMIC-CXR \cite{physionet,mimic,mimic-jpg} and CheXpert \cite{chexpert,chexpert-plus}. However, model reliability and performance remain constrained by incomplete coverage of clinically meaningful patterns in available training data \cite{aug1,aug2,aug3,aug_medical}. While dataset labels capture high-level diagnostic categories, many important configurations of underlying radiographic findings—particularly rare, subtle, or co-occurring patterns—are sparsely represented or entirely absent, limiting generalisation and increasing uncertainty in real-world settings. In this work, we distinguish between \emph{pathologies}, which denote high-level diagnostic labels (e.g., \textit{Suspicious Malignancy}), and \emph{clinical concepts}, the specific radiographic findings that support those diagnoses (e.g., lung mass, adenopathy). Although pathology-level label diversity may appear high, incomplete coverage of clinically meaningful \emph{concept combinations} remains a fundamental limitation.

\begin{figure}[t]
    \centering
    \includegraphics[width=0.8\linewidth]{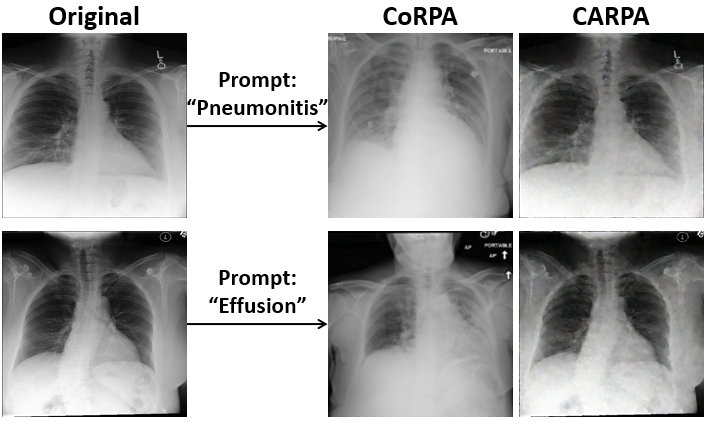}
    \caption{Example synthetic images generated by CoRPA and CARPA. Image generation prompts: (1) Pneumonitis, (2) Effusion}
    \label{fig:examples}
    \vspace{-5pt}
\end{figure}

To mitigate data scarcity and improve generalisation, data augmentation has long been a standard component of medical image analysis pipelines. Conventional computer vision–based transformations (e.g., rotation, flipping, intensity perturbations) are routinely applied and have been shown to improve robustness and reduce overfitting across a wide range of medical imaging tasks, including CXR classification \cite{aug_medical,data_aug}. These techniques increase sample diversity within the support of the original data distribution and are widely adopted as baseline augmentation strategies. 
Synthetic image generation offers a promising mechanism for expanding training distributions beyond observed data. Generative Adversarial Networks (GANs) \cite{gans} and diffusion-based approaches \cite{diffusion} have explored image-level synthesis and augmentation for CXRs, improving downstream model performance, particularly in low-data or class-imbalanced datasets \cite{aug_medical_improve}. However, most existing synthesis and augmentation approaches operate at the image level and do not explicitly reason about clinical semantics. As a result, increased visual diversity does not necessarily translate into improved coverage of clinically meaningful concept combinations, and synthetic images may remain redundant or clinically ambiguous \cite{gan_apply3}.
CoRPA \cite{corpa} is a recently proposed concept-based radiographic perturbation framework that explicitly manipulates clinical concept representations to generate synthetic CXRs reflecting targeted diagnostic variations. By operating in a structured clinical concept space, CoRPA demonstrated that concept-level perturbations can more systematically explore underrepresented clinical scenarios than purely image-driven approaches. However, CoRPA does not enforce anatomical constraints (e.g., lung shape, skeletal structure) between original and synthetic images, which leads to structural inconsistencies and limits the clinical utility of generated data. Figure \ref{fig:examples} shows two examples of CoRPA-generated CXRs with completely different chest structures than their original images.

We introduce \textbf{CARPA} (Clinically Aware Radiographic Perturbation Augmentation), a clinically and anatomically grounded refinement of concept-based synthetic CXR generation. CARPA preserves anatomical structure by performing targeted concept perturbations through image editing operations, enabling controlled modification of pathological findings while maintaining radiographic plausibility. By enforcing anatomical correspondence and introducing a refined set of clinically meaningful perturbation operations, CARPA aims to expand concept space coverage in a manner that directly benefits downstream learning.
We evaluate CARPA across seven convolutional backbone architectures by fine-tuning models on stratified subsets of synthetic images and testing on a held-out MIMIC-CXR benchmark. We analyze changes in multi-label classification performance, predictive uncertainty, and calibration after fine-tuning with both CoRPA- and CARPA-generated data. We also assess structural fidelity using image similarity metrics, semantic concept alignment using a vision–language model \cite{llavarad,uncertainty_pipeline}, and conduct an expert evaluation in which two radiologists assess realism and agreement with proposed clinical findings. CARPA consistently outperforms CoRPA, yielding stronger model performance gains, lower predictive and semantic uncertainty, and substantially higher anatomical fidelity to original images. Expert assessment further indicates that CARPA images are more clinically plausible, with disagreements primarily reflecting realistic diagnostic ambiguity rather than structural or semantic errors. Together, these results show that anatomically grounded concept perturbations provide a more effective and clinically meaningful approach to synthetic data augmentation for CXR models.

\section{Background and Related Work}

\subsection{Medical Image Augmentation and Synthesis}

Data augmentation is a long-standing strategy for improving deep learning performance in medical image analysis. Classical computer vision augmentations (e.g., rotation, flipping, intensity perturbations) are widely used and have demonstrated consistent benefits for chest X-ray classification tasks \cite{aug_medical,aug1,aug3}. However, these methods operate strictly within the support of the original image distribution and do not introduce new clinical semantics or diagnostic scenarios.
Synthetic image generation addresses this limitation by expanding training distributions beyond observed data. Generative Adversarial Networks (GANs) \cite{gans} have been applied to chest X-ray synthesis for data augmentation and class imbalance mitigation, with several studies reporting performance gains in limited-data settings \cite{gans_apply,gans_apply2}. Diffusion-based generative models \cite{diffusion} and LLMs \cite{llmcxr} have also demonstrated improved image fidelity and stability and have been adopted for medical image synthesis \cite{synth1,synth2,new}. Despite producing visually realistic images, most generative approaches optimise image-level realism without explicit control over clinical findings. As a result, synthetic images may increase visual diversity without meaningfully expanding coverage of clinically relevant concept combinations \cite{gan_apply3,gancoverage}.

\subsection{Concept-Driven Synthetic Image Generation}

To address the limitations of image-driven synthesis, recent work explores incorporating structured clinical information into the generation process. Concept-based representations derived from radiology reports or expert-defined vocabularies provide a natural abstraction for modelling diagnostic findings and their interactions. Operating at the concept level enables targeted manipulation of clinically meaningful attributes and supports interpretable analysis of model behaviour.
CoRPA \cite{corpa} represents a notable advance in this direction. CoRPA encodes radiology reports as binary clinical concept vectors using an expert-defined vocabulary and NLP-based extraction, and generates synthetic chest X-rays by perturbing these vectors and using them as prompts within a report-to-image generation pipeline. The framework defines two perturbation strategies: intra-class perturbations, which modify concepts within a diagnostic category, and outer-class perturbations, which introduce cross-diagnostic changes. This design enables systematic exploration of underrepresented regions of the clinical concept space and facilitates targeted stress-testing of model behaviour.
However, CoRPA does not preserve anatomical correspondence between original and synthetic images (Figure \ref{fig:examples}). Because synthetic images are generated solely from perturbed reports, the resulting samples may differ substantially in anatomical structure (e.g., lung shape, skeletal structure) and global appearance from their source images. While suitable for adversarial or out-of-distribution evaluation, this lack of anatomical grounding can limit clinical interpretability and reduce the effectiveness of synthetic images for improving model performance and reliability.

\subsection{Predictive Uncertainty and Calibration}

Reliable deployment of medical imaging models requires not only high classification accuracy but also trustworthy estimates of predictive confidence, as model outputs often inform high-stakes clinical decisions \cite{unc1,unc2}. Predictive uncertainty is commonly quantified using entropy-based measures derived from output probabilities. In multi-label settings, predictive entropy reflects the degree of ambiguity in a model’s predictions, with higher values indicating uncertainty arising from borderline evidence, noise, or distributional shift \cite{entropy}.
Calibration metrics assess a complementary aspect of reliability: the alignment between predicted probabilities and empirical correctness. Expected Calibration Error (ECE) is widely used to quantify whether predicted confidences reflect true outcome frequencies \cite{ece}. Well-calibrated models assign probabilities that accurately correspond to observed risk, while poorly calibrated models may exhibit systematic overconfidence even when performance is high. In medical imaging, miscalibration is particularly problematic, as it undermines trust in probabilistic outputs and downstream decision thresholds \cite{calibration,calibration2}. Bayesian approaches, such as Monte Carlo dropout and deep ensembles, estimate epistemic uncertainty by modelling uncertainty in learned parameters, but they are not uniformly supported across architectures and introduce additional computational complexity. 

\subsection{Semantic Uncertainty and Vision–LLMs}

We also emphasize the importance of evaluating uncertainty at the semantic level, particularly for structured clinical interpretation tasks \cite{semantic}. In medical imaging, errors often arise not only from low-confidence predictions, but from semantically incorrect or incomplete clinical findings. Semantic uncertainty therefore characterizes discrepancies between the clinical concepts intended or present in an image and those inferred by a model \cite{semanticrisk}.
Vision–language models provide a natural mechanism for extracting structured semantic information from medical images. Models such as BioViL \cite{biovilt} and GLoRIA \cite{gloria} learn joint image–text representations from paired radiology reports, while approaches such as LLaVA-Rad \cite{llavarad} extend large-scale image–language architectures to radiological image-to-text generation. These models enable flexible extraction of clinically meaningful semantics beyond fixed label sets.

\cite{uncertainty_pipeline} recently proposed a semantic uncertainty quantification pipeline that formalizes this idea by comparing predicted and ground-truth concept sets using set-based metrics. Model outputs are mapped to structured concept representations and evaluated using precision, recall, and similarity measures, enabling decomposition of uncertainty into over-prediction (hallucinated findings) and under-prediction (missed findings). This aligns uncertainty estimates with clinically meaningful failure modes and provides finer-grained insight than aggregate accuracy metrics.
In this work, we adapt this semantic uncertainty framework using an image-to-text model. CARPA’s synthetic chest X-rays are processed with LLaVA-Rad to generate free-text descriptions, which are converted into clinical concept vectors and compared to the ground-truth concept perturbations used during image generation. This enables evaluation of not only visual realism, but also whether synthetic images convey the intended clinical semantics, providing a principled semantic uncertainty assessment for concept-driven image synthesis.

\section{Materials and Experimental Setup}

\subsection{Dataset}
\begin{figure}[t]
    \centering

    \begin{subfigure}[b]{\linewidth}
        \centering
        \includegraphics[width=0.95\linewidth]{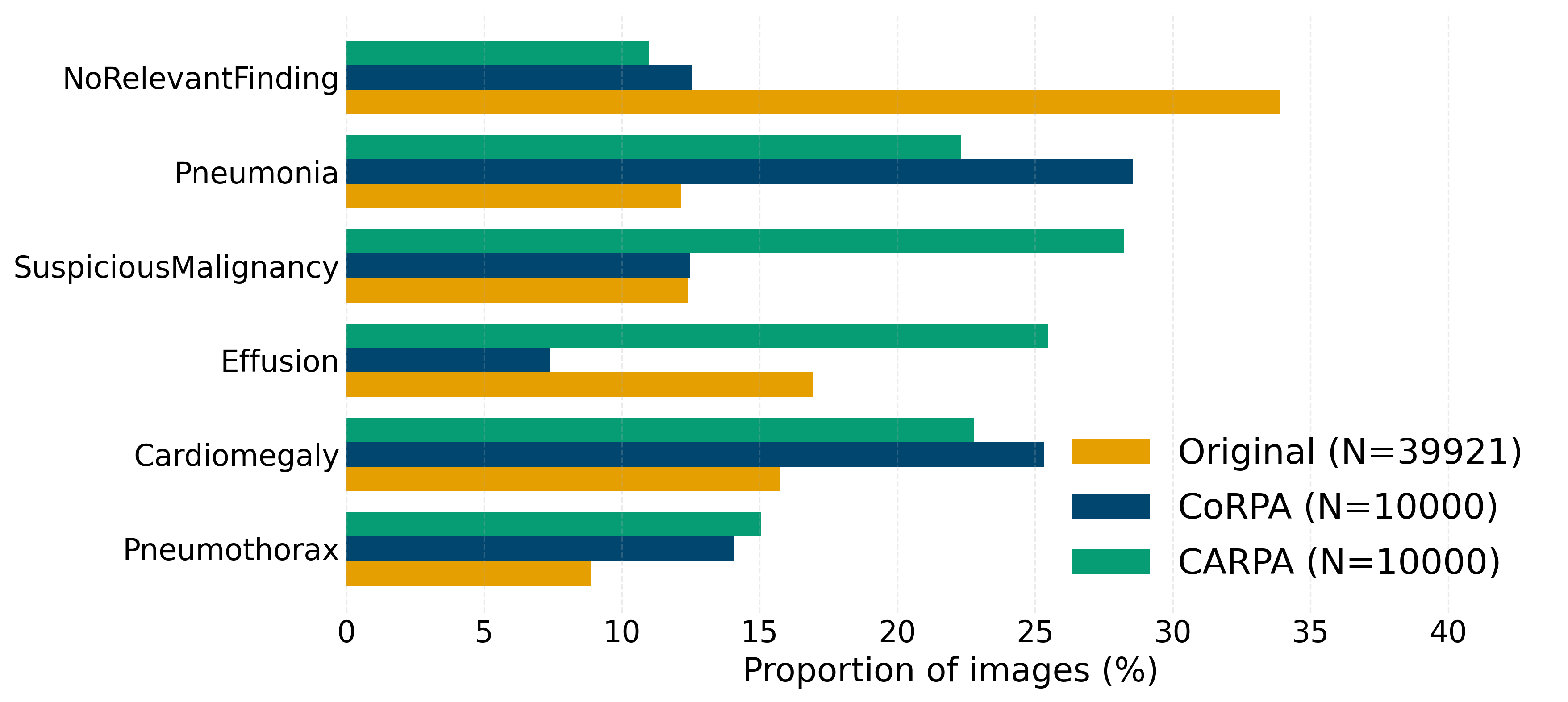}
        \caption{Label distribution}
        \label{fig:labels}
    \end{subfigure}
    \begin{subfigure}[b]{\linewidth}
        \centering
        \includegraphics[width=\linewidth]{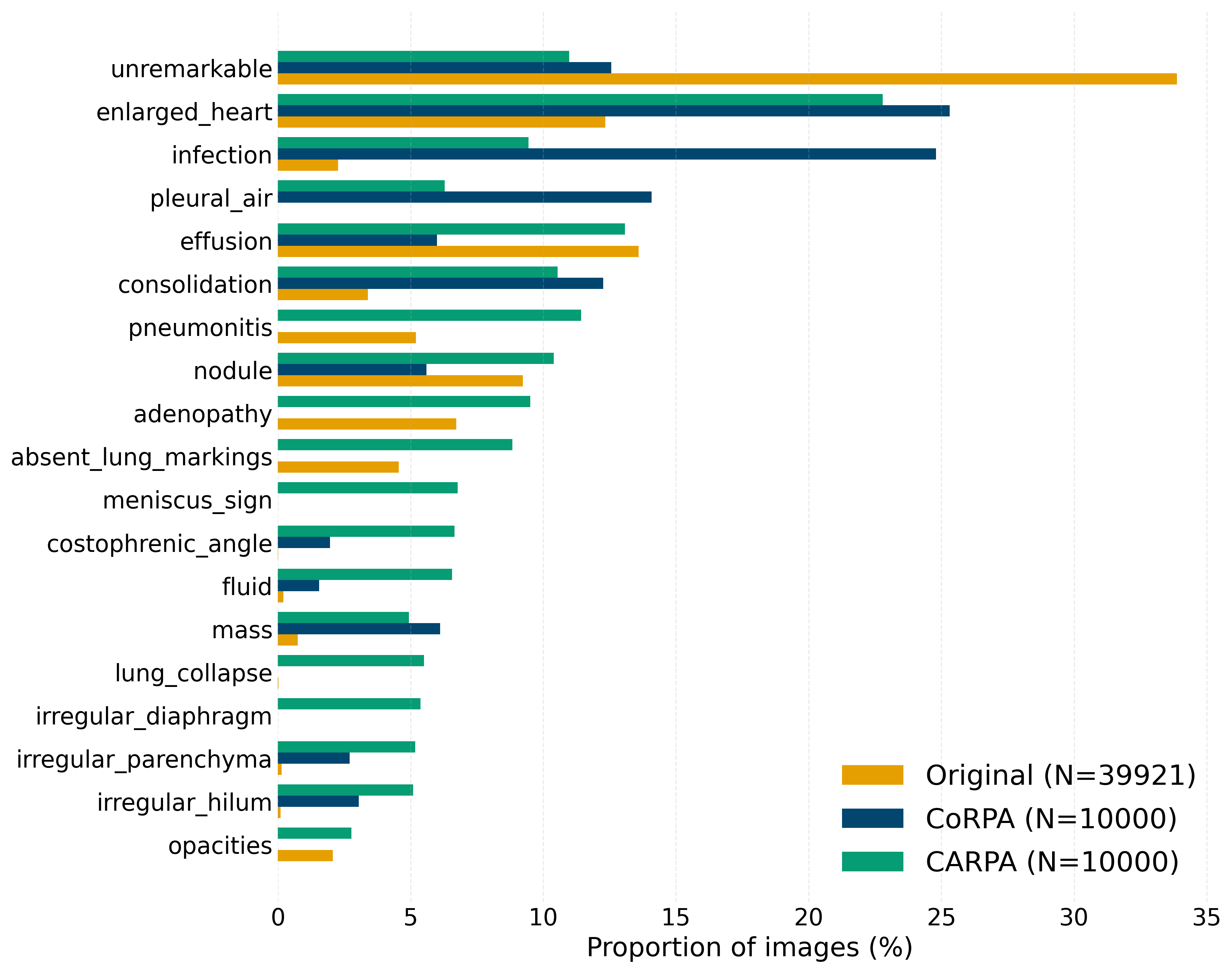}
        \caption{Concept distribution}
        \label{fig:concepts}
    \end{subfigure}

    \caption{Comparison of label (a) and concept (b) distributions between the original dataset (N=39921), CoRPA (N=10000) and CARPA (N=10000). Values are proportions of respective datasets.}
    \label{fig:concept_space}
\end{figure}

We use the MIMIC-CXR dataset \cite{mimic,mimic-jpg,physionet}, a large public collection of chest radiographs paired with free-text radiology reports. We restrict the dataset to posteroanterior (PA) views to reduce variability in imaging geometry and interpretation. Following CoRPA \cite{corpa}, we do not use the provided dataset labels, which have been shown to be noisy and unreliable \cite{unreliable}. Instead, we re-annotate all images using an expert-defined clinical concept space. Under radiologist guidance, we focus on six clinically relevant diagnostic labels: \textbf{Pneumothorax} (features concerning for pleural air), \textbf{Pneumonia} (airspace disease consistent with infection), \textbf{Pleural Effusion} (pleural fluid or effusion-related findings), \textbf{Cardiomegaly} (enlarged cardiac silhouette), \textbf{Suspicious Malignancy} (features warranting lung cancer workup), and \textbf{No Relevant Finding} (absence of relevant abnormalities). The original label distribution is highly imbalanced, with a predominance of normal studies. To mitigate this, we randomly undersample the \textbf{No Relevant Finding} class to twice the size of the largest abnormal class. The resulting dataset contains 40,083 unique images, each of which may have multiple labels. Class counts are shown in Figure~\ref{fig:concept_space}. We perform a multi-label stratified split into training (80\%), validation (10\%), and test (10\%) sets. The held-out test set contains 4,008 images and is used consistently for all experiments.

\subsubsection{Clinical Concept Annotation}\label{annotation}
Each image is annotated by converting its associated free-text radiology report into a binary clinical concept vector, following the CoRPA framework. Our concept space is based on the CoRPA vocabulary, with minor refinements introduced under expert guidance. Radiology reports are converted into binary vectors indicating the presence or absence of these predefined clinical concepts using expert-defined phrase mappings. Images may contain multiple pathology concepts; when any pathology is present, the \textit{Unremarkable} concept is suppressed. Clinical concepts are deterministically mapped to diagnostic labels, ensuring consistent annotation across real and synthetic datasets and enabling direct comparison in the concept space. The full concept vocabulary, phrase mappings, and label definitions are publicly available.

\subsection{Model Architectures}
We train and evaluate seven convolutional backbone architectures, matching those used in CoRPA and the MICCAI CXR-LT challenge \cite{cxr-lt}: ConvNeXt-Base, ConvNeXt-Small, DenseNet-161, EfficientNet-V2-S, ResNet-50, ResNet-101, and ResNeXt-101. All models are initialized with ImageNet weights and adapted for multi-label classification.
Models are trained on the MIMIC-CXR training set using a multi-label formulation, as radiographs may exhibit multiple co-occurring findings. Labels are encoded as multi-hot vectors; the \textbf{No Relevant Finding} label is treated implicitly as the absence of positive pathology labels rather than as a competing class. Input images are normalized using ImageNet statistics. Optimization is performed using AdamW, and we use binary cross-entropy loss and per-class positive weights computed from the training set to address class imbalance. Models are trained for 30 epochs with batch size 32 and learning rate $3\times10^{-4}$. Final performance is reported using macro-averaged AUROC, AUPRC, and F1, with per-class thresholds tuned on the validation set.

\subsubsection{Fine-Tuning on Synthetic Data}
To assess the impact of synthetic concept perturbations, each trained model is fine-tuned on stratified subsets of synthetic images generated using either CoRPA or CARPA. As the two pipelines produce different numbers of images, we uniformly sample 10,000 images from each to ensure comparability. Synthetic images are annotated using the same concept-to-label mapping as real data; samples mapping to empty label sets are discarded. Each synthetic subset is split into training and validation partitions (80/20). Fine-tuning is performed for five epochs using head-only adaptation - only the classification head is optimised. We use AdamW with a low learning rate and binary cross-entropy loss with logits. Final evaluation is performed on the original held-out MIMIC-CXR test set to ensure direct comparability with baseline results.

\subsection{Model Performance and Uncertainty}
For baseline, CoRPA-finetuned, and CARPA-finetuned models, we evaluate downstream performance using macro-averaged AUROC, AUPRC, and F1. AUROC measures discriminative ability across thresholds and is insensitive to class imbalance, while AUPRC emphasizes positive-class performance under imbalance. Macro-F1 summarizes precision–recall trade-offs with equal weight across labels. To assess model reliability, we evaluate predictive uncertainty and calibration using entropy and Expected Calibration Error (ECE), computed directly from model output probabilities. Predictive entropy reflects confidence in multi-label predictions, while ECE measures alignment between predicted probabilities and empirical correctness. These metrics are architecture-agnostic and applicable to all evaluated backbones. Lower entropy indicates more confident predictions, while ECE values closer to zero indicate better calibration. Together, these measures provide a complementary view of model reliability beyond discrimination performance alone.

\subsection{Synthetic Image and Semantic Evaluation}
We also evaluate the quality and content of synthetic images produced by CoRPA and CARPA. Structural fidelity is assessed using the Structural Similarity Index (SSIM) between original–synthetic image pairs, reported as mean $\pm$ standard deviation. SSIM measures preservation of anatomical structure and is well suited for radiographic consistency analysis.
Semantic uncertainty is evaluated using a concept-level pipeline adapted from \cite{uncertainty_pipeline}. Synthetic images are processed using LLaVA-Rad \cite{llavarad}, which generates free-text descriptions that are converted into predicted clinical concept vectors. 
For each image, we measure uncertainty in the clinical concept space by directly comparing the ground-truth and predicted concept vectors. 
Treating these vectors as sets, we compute Jaccard similarity and normalized Hamming distance. We further define two uncertainty measures: $U_{\mathrm{FP}} = 1-\mathrm{Precision}$, capturing over-predicted or hallucinated findings, and $U_{\mathrm{FN}} = 1-\mathrm{Recall}$, capturing missed findings.

\subsection{Expert Evaluation}\label{expert}
Finally, we conduct an expert evaluation involving two independent radiologists, who assess subsets of 50 CoRPA and 100 CARPA synthetic images. Radiologists evaluate each image for perceived realism (real vs. synthetic) and clinical concordance with the proposed concept annotations (fully agree, partially agree, disagree), and provided free-text explanations when images were judged synthetic or when concept disagreement occurred. We report response distributions and inter-rater agreement to provide qualitative validation of the synthetic data and concept design.

\section{CARPA - Clinically Aware Radiographic Perturbation Augmentation}

CARPA is designed to expand clinically meaningful concept coverage in chest X-ray datasets while preserving anatomical correspondence to real radiographs. It generates anatomically faithful synthetic images by applying structured perturbations in a clinical concept space and translating these perturbations into localized image edits on an original scan. CARPA consists of three stages. First, each image–report pair is encoded as a binary clinical concept vector indicating the presence or absence of expert-defined radiological findings, following the report-derived representation used in CoRPA \cite{corpa} (full concept lists are provided in our GitHub repository). Second, the concept vector is modified using clinically motivated perturbation strategies—\emph{Intra-Class}, \emph{Insertion}, and \emph{Deletion}—designed to introduce under-represented yet plausible combinations of findings. We note that this represents an improved perturbation set when compared to CoRPA, which used simplified Intra-Class and Outer-Class perturbations. Third, the perturbed concept vector is converted into a concise clinical prompt and passed to RadEdit \cite{radedit}, which applies spatially localized edits to the original image to express the intended concept changes while preserving global anatomical structure.
By operating directly in the clinical concept space and enforcing anatomical grounding at the image level, CARPA decouples semantic expansion from uncontrolled structural variation. This design enables systematic exploration of clinically relevant concept combinations while maintaining radiographic plausibility, producing synthetic images that are well suited for improving model performance, calibration, and uncertainty behaviour.

\begin{table*}[t]
\centering
\small
\begin{tabular}{lrrrrrrrr}
\toprule
& & ConvNeXt-B & ConvNeXt-S & DenseNet & Efficient & ResNet50 & ResNet101 & ResNeXt101 \\
\midrule
Baseline & AUROC & 0.947 & 0.940 & 0.922 & 0.943 & 0.913 & 0.926 & 0.949 \\
& AUPRC & 0.864 & 0.838 & 0.772 & 0.839 & 0.719 & 0.786 & 0.869 \\
& F1 & 0.835 & 0.802 & 0.734 & 0.797 & 0.677 & 0.740 & 0.844 \\
\midrule
Finetuned CARPA & $\Delta$AUROC & $\downarrow$ 0.007 & $\downarrow$ 0.008 & $\downarrow$ 0.008 & $\downarrow$ 0.010 & \textbf{$\uparrow$ 0.003} & $\downarrow$ 0.007 & $\downarrow$ 0.009 \\
(Ours) & $\Delta$AUPRC & \textbf{$\uparrow$ 0.048} & \textbf{$\uparrow$ 0.052} & \textbf{$\uparrow$ 0.110} & \textbf{$\uparrow$ 0.026} & \textbf{$\uparrow$ 0.087} & \textbf{$\uparrow$ 0.058} & $\downarrow$ 0.003 \\
& $\Delta$F1 & \textbf{$\uparrow$ 0.087} & \textbf{$\uparrow$ 0.092} & \textbf{$\uparrow$ 0.096} & \textbf{$\uparrow$ 0.019} & \textbf{$\uparrow$ 0.102} & \textbf{$\uparrow$ 0.033} & \textbf{$\uparrow$ 0.002} \\
\midrule
Finetuned CoRPA & $\Delta$AUROC & $\downarrow$ 0.015 & $\downarrow$ 0.016 & $\downarrow$ 0.014 & $\downarrow$ 0.012 & $\downarrow$ 0.021 & $\downarrow$ 0.019 & $\downarrow$ 0.008 \\
& $\Delta$AUPRC & $\downarrow$ 0.061 & $\downarrow$ 0.058 & $\downarrow$ 0.074 & $\downarrow$ 0.055 & $\downarrow$ 0.077 & $\downarrow$ 0.082 & $\downarrow$ 0.036 \\
& $\Delta$F1 & $\downarrow$ 0.137 & $\downarrow$ 0.131 & $\downarrow$ 0.136 & $\downarrow$ 0.077 & $\downarrow$ 0.129 & $\downarrow$ 0.155 & $\downarrow$ 0.102 \\
\bottomrule
\end{tabular}
\caption{Comparison of multi-label classification performance across seven backbone architectures. We report macro-averaged AUROC, AUPRC, and F1 for baseline models trained on a subset of MIMIC-CXR, and performance deltas (increase $\uparrow$ or decrease $\downarrow$) for corresponding variants fine-tuned on the CARPA (Ours) or CoRPA synthetic datasets. All metrics are evaluated on the same held-out test split (N=4008), with class-specific decision thresholds calibrated on the validation set when computing F1. Performance increases shown in \textbf{bold}.}
\label{tab:classification_acc}
\end{table*}

\begin{table*}[t]
\centering
\small
\begin{tabular}{lrrrrrrrr}
\toprule
& & ConvNeXt-B & ConvNeXt-S & DenseNet & Efficient & ResNet50 & ResNet101 & ResNeXt101 \\
\midrule
Baseline & Entropy & 0.086 & 0.135 & 0.225 & 0.123 & 0.265 & 0.201 & 0.043 \\
& ECE & 0.045 & 0.063 & 0.093 & 0.062 & 0.114 & 0.085 & 0.040 \\
\midrule
Finetuned CARPA & $\Delta$Entropy & \textbf{$\downarrow$ 0.012} & \textbf{$\downarrow$ 0.022} & $\uparrow$ 0.026 & $\uparrow$ 0.024 & \textbf{$\downarrow$ 0.105} & \textbf{$\downarrow$ 0.075} & \textbf{$\downarrow$ 0.002} \\
(Ours) & $\Delta$ECE & \textbf{$\downarrow$ 0.012} & \textbf{$\downarrow$ 0.008} & \textbf{$\downarrow$ 0.014} & \textbf{$\downarrow$ 0.010} & \textbf{$\downarrow$ 0.031} & $\uparrow$ 0.006 & \textbf{$\downarrow$ 0.009} \\
\midrule
Finetuned CoRPA & $\Delta$Entropy & $\uparrow$ 0.057 & $\uparrow$ 0.055 & $\uparrow$ 0.062 &$\uparrow$ 0.071 & $\uparrow$ 0.032 & $\uparrow$ 0.035 & $\uparrow$ 0.062 \\
& $\Delta$ECE & $\uparrow$ 0.094 & $\uparrow$ 0.097 & $\uparrow$ 0.080 & $\uparrow$ 0.125 & $\uparrow$ 0.051 & $\uparrow$ 0.083 & $\uparrow$ 0.154\\
\bottomrule
\end{tabular}
\caption{Uncertainty and calibration analysis across seven backbone architectures. For each model, we report predictive entropy and Expected Calibration Error (ECE) for the baseline model trained on the MIMIC-CXR subset, as well as performance deltas (increase $\uparrow$ or decrease $\downarrow$) for variants fine-tuned on the CARPA (Ours) and CoRPA synthetic datasets. All metrics are computed on the same held-out test split (N=4008), with values reported as macro-averages across labels. Uncertainty reductions shown in \textbf{bold}.}
\label{tab:model_uncertainty}
\end{table*}

\subsection{Concept Perturbation Types}
Let $V \in \{0,1\}^C$ denote the original clinical concept vector for an image, and $L$ denote its associated diagnostic label set induced by concept-to-label mapping (Section~\ref{annotation}). CARPA produces perturbed vectors $V'$ using three perturbation types:

\begin{itemize}
    \item \textbf{Intra-Class.} Modifies concepts associated with the ground-truth diagnostic label to reflect alternative but valid manifestations of the same condition, capturing within-diagnosis variability. 

    \item \textbf{Insertion.} Adds concepts from an additional pathology label $L'$ while retaining the original label(s) $L$, simulating multi-pathology presentations and spurious findings.

    \item \textbf{Deletion.} Removes concepts associated with a selected pathology. In multi-label cases, one label is randomly selected for removal; in single-label cases, concepts are replaced with the \textit{No Relevant Finding} (unremarkable) concept, simulating a clinically plausible missed diagnosis.
\end{itemize}

Intra-class variation is not defined for single-concept pathology labels (e.g., \textbf{Cardiomegaly} has only one concept \textbf{Enlarged Heart}), where meaningful substitutions are not possible. For deletion, single-label abnormal cases are mapped to \textbf{No Relevant Finding}. We generate up to two unique perturbations per variation type for each image, except where constrained (e.g., single-label deletion).

\subsection{Anatomy-Preserving Image Editing}
Each perturbed concept vector is converted into a concise clinical prompt listing the active concepts. We do not use explicit ``remove'' phrasing; instead, RadEdit is conditioned on the target findings to be present in the edited image. The prompt guides RadEdit to generate an edited radiograph $x'$ from the original image $x$, preserving anatomical correspondence while expressing the intended diagnostic change.
RadEdit combines a latent diffusion model \cite{latent}, the BioViL-T text encoder \cite{biovilt}, and the SDXL-VAE autoencoder \cite{sdxl} to apply fine-grained, spatially localized edits guided by clinical text. Unlike full-image synthesis methods, RadEdit edits the source radiograph directly, encouraging preservation of global structure (e.g., lung fields and bony anatomy) while enabling localized diagnostic changes. Because RadEdit was trained on MIMIC-CXR using radiology report language aligned with our annotation pipeline, it interprets our concept-derived prompts with high fidelity, producing radiologically coherent and anatomically consistent outputs.

\section{Results}

\begin{table}[t]
\centering
\small
\begin{tabular}{lrr}
\toprule
 & Perturbation Type & SSIM ($Mean \pm sd$) \\
\midrule
CARPA  & Intra-class      & $0.748 \pm 0.030$ \\
  (Ours) & Insertion        & $0.751 \pm 0.032$ \\
  & Deletion         & $0.749 \pm 0.030$ \\
  & \textbf{Overall} & $\mathbf{0.750 \pm 0.031}$ \\
\midrule
CoRPA & Intra-class  & $0.304 \pm 0.063$ \\
 & Outer-class  & $0.229 \pm 0.102$ \\
 & \textbf{Overall} & $\mathbf{0.273 \pm 0.089}$ \\
\bottomrule
\end{tabular}
\caption{Structural similarity (SSIM) between original chest X-rays and synthetic images generated by CARPA (ours) and CoRPA. Higher SSIM indicates greater preservation of anatomical structure.
}
\label{tab:ssims}
\end{table}

\begin{table*}[h]
\small
\centering
\begin{tabular}{lrrrr}
\toprule
 & Jaccard & Hamming & $U_{\text{FP}}{=}1{-}\text{Precision}$ & $U_{\text{FN}}{=}1{-}\text{Recall}$ \\
\midrule
CARPA (Ours)   & $0.609 \pm 0.149$ & $0.092 \pm 0.043$ & $0.489 \pm 0.211$ & $0.287 \pm 0.114$ \\
CoRPA & $0.168 \pm 0.327$ & $0.113 \pm 0.068$ & $0.799 \pm 0.359$ & $0.595 \pm 0.488$ \\
\bottomrule
\end{tabular}
\caption{Semantic concept uncertainty evaluation for CARPA (Ours) and CoRPA subsets using the LLaVA-Rad image-to-text pipeline. Values are reported as mean $\pm$ standard deviation over 10,000 images.}
\label{tab:llavarad_semantic_uncertainty}
\end{table*}

\begin{table}[h]
\small
\centering
\begin{tabular}{lrrrr}
\toprule
 & Task & Response & R1 & R2  \\
\midrule
CARPA  
 & Real / Synthetic & Real & \textbf{94\%} & \textbf{88\%}  \\
(Ours) &  & Synthetic & 6\% & 12\%  \\
 & Clin. Agreement & Fully Agree & \textbf{61\%} & \textbf{59\%} \\
 &  & Partial Agree & 22\% & 21\% \\
 &  & Disagree & 17\% & 20\%  \\
\midrule
CoRPA  
 & Real / Synthetic & Real & 86\% & 62\%  \\
 &  & Synthetic & 14\% & 38\% \\
 & Clin. Agreement & Fully Agree & 46\% & 52\%  \\
 &  & Partial Agree & \textbf{24\%} & \textbf{24\%} \\
 &  & Disagree & 30\% & 24\%  \\
\bottomrule
\end{tabular}
\caption{Expert evaluation of synthetic chest X-rays generated by CARPA (N=100) and CoRPA (N=50). Two radiologists (R1, R2) independently assessed images for realism (Real / Synthetic)
and agreement with proposed clinical findings (Clin. Agreement).}
\label{tab:expert_agreement}
\end{table}

\subsection{Downstream Classification Performance}
Table~\ref{tab:classification_acc} summarizes multi-label classification performance across seven backbone architectures. Baseline models trained on MIMIC-CXR achieve strong macro-AUROC and AUPRC scores. Fine-tuning on CARPA-generated synthetic images leads to consistent improvements in macro-AUPRC and macro-F1 across all architectures, with gains particularly pronounced for DenseNet-161, ResNet-50, and ConvNeXt variants. In contrast, fine-tuning on CoRPA-generated images consistently degrades performance across all three metrics. While AUROC changes after CARPA fine-tuning are modest and occasionally negative, AUPRC and F1 show systematic improvements, indicating better precision–recall trade-offs under class imbalance. These results suggest that CARPA-generated synthetic images improve decision quality for positive findings, even when overall ranking performance remains stable. The consistent performance degradation observed for CoRPA indicates that unconstrained concept perturbations may introduce noise that hinders learning.

\subsection{Predictive Uncertainty and Calibration}
Table~\ref{tab:model_uncertainty} reports changes in predictive entropy and Expected Calibration Error (ECE) following fine-tuning. Fine-tuning on CARPA leads to reductions in predictive entropy for most architectures, notably for ResNet-50, ResNet-101, and ConvNeXt models, indicating increased confidence in predictions. Calibration also improves consistently, with ECE decreasing for six of seven architectures. In contrast, CoRPA fine-tuning increases both entropy and ECE across all architectures, reflecting higher uncertainty and degraded probability calibration. These results indicate that CARPA improves not only classification performance but also predictive reliability, while CoRPA increases uncertainty despite increasing visual diversity. The divergence between CARPA and CoRPA highlights the importance of anatomical and semantic grounding when using synthetic data for model refinement.

\subsection{Structural Fidelity of Synthetic Images}
We evaluate the anatomical consistency of synthetic images using Structural Similarity Index (SSIM), reported in Table~\ref{tab:ssims}. CARPA-generated images achieve substantially higher SSIM scores than CoRPA across all perturbation types, with an overall mean SSIM of $0.750 \pm 0.031$ compared to $0.273 \pm 0.089$ for CoRPA. This indicates that CARPA preserves anatomical structure much more faithfully, consistent with its image-editing-based perturbation strategy. Within CARPA, intra-class, insertion, and deletion perturbations yield similar SSIM values, suggesting that targeted concept insertions and removals can be performed without disrupting global image structure. CoRPA’s outer-class perturbations show particularly low SSIM, reflecting substantial structural divergence from the source images. These results confirm that CARPA maintains anatomical correspondence while expanding the clinical concept space. Examples are shown in Figure \ref{fig:examples}.

\subsection{Semantic Concept Uncertainty}
Table~\ref{tab:llavarad_semantic_uncertainty} reports semantic agreement between ground-truth concept vectors and the clinical concepts inferred by LLaVA-Rad. For CoRPA, we observe low average semantic overlap, with a mean Jaccard similarity of $0.168 \pm 0.327$ and a relatively high Hamming distance of $0.113 \pm 0.068$. CoRPA exhibits high over-prediction uncertainty ($U_{\mathrm{FP}} = 0.799 \pm 0.359$), indicating frequent hallucination of unsupported clinical findings, as well as high under-prediction uncertainty ($U_{\mathrm{FN}} = 0.595 \pm 0.488$), reflecting missed ground-truth concepts. This suggests unstable semantic grounding, with different images dominated by different failure modes. From a clinical perspective, this implies that concept extraction on CoRPA images is unreliable in both directions, introducing extraneous findings while failing to consistently recover intended abnormalities.
In contrast, CARPA demonstrates markedly improved semantic consistency. Jaccard similarity increases to $0.609 \pm 0.149$, while Hamming distance decreases to $0.092 \pm 0.043$, indicating stronger and more stable concept alignment. Both over-prediction and under-prediction uncertainty are substantially reduced ($U_{\mathrm{FP}} = 0.489 \pm 0.211$, $U_{\mathrm{FN}} = 0.287 \pm 0.114$), with lower variance across images. These results indicate that CARPA-generated images convey intended clinical concepts more consistently than CoRPA-generated images.

\subsection{Expert Evaluation and Qualitative Analysis}
From our expert evaluation (Section \ref{expert}), results are summarized in Table~\ref{tab:expert_agreement}, with additional insight provided by qualitative analysis of radiologists’ free-text responses. Radiologists, \texttt{R1} and \texttt{R2}, judged CARPA images as real at substantially higher rates than CoRPA images (\texttt{R1}: 94\% vs.\ 86\%, \texttt{R2}: 88\% vs.\ 62\%), and reported higher full clinical agreement with the proposed concept annotations. Inter-rater agreement was moderate to strong, with radiologists agreeing on realism judgments for 61\% of images and on clinical concept judgements for 72\%.
Qualitative analysis reveals clear differences in the nature of uncertainty between CoRPA and CARPA images. For CoRPA, radiologists most frequently cited anatomical inconsistencies as the primary reason images appeared synthetic. Common issues included implausible rib spacing, asymmetric clavicles, distorted cardiac silhouettes, and non-physiological lung textures. These images were often described as “almost right” but failing under closer inspection, suggesting subtle structural errors rather than visual artifacts. Clinical disagreement for CoRPA was most commonly radiologists agreeing that an abnormality was present but disputing whether all proposed concepts were supported, as well as over-specified findings and insufficiently clear hallmark signs (e.g., pleural lines or meniscus signs).
In contrast, for CARPA, major anatomical errors were rarely noted; instead, images judged synthetic were typically described as “too clean,” or exhibiting subtle textural inconsistencies. Disagreements regarding concepts were less frequent and narrower in scope, usually concerning secondary or borderline findings rather than the primary diagnosis. Radiologists often used language such as “subtle,” “early,” or “not definitive,” indicating uncertainty arising from inherent image ambiguity rather than incorrect or hallucinated findings.
Overall, CARPA reduces gross semantic and anatomical failure and shifts uncertainty toward clinically realistic ambiguity. While CoRPA assessments often reflected structural implausibility or concept over-specification, CARPA assessments more closely resemble the uncertainty encountered in real-world radiographic interpretation. This transformation of uncertainty provides strong human evidence that anatomical grounding improves not only realism but also the clinical significance of uncertainty in synthetic chest X-rays.

\section{Conclusion}

We introduced CARPA, a clinically aware and anatomically grounded refinement of concept-based radiographic perturbation for synthetic chest X-ray generation. By performing targeted concept perturbations via anatomy-preserving image editing, CARPA expands clinical concept coverage while maintaining radiographic plausibility. Across seven backbone architectures, fine-tuning on CARPA-generated images consistently improved multi-label classification performance while reducing predictive uncertainty and improving calibration.
CARPA also produced synthetic images with high structural fidelity to their source images, and strong semantic alignment between intended and inferred clinical concepts. Concept-level uncertainty analysis using a vision–language model showed reductions in both over- and under-prediction, when compared to prior CoRPA work, shifting errors from gross semantic mismatches towards clinically realistic ambiguity. Expert evaluation further supported these findings, with radiologists judging CARPA images as largely realistic and demonstrating high clinical agreement, with disagreements primarily reflecting borderline or subtle findings rather than structural implausibility. 

\paragraph{Code.} https://anonymous.4open.science/r/CARPA-3BAF/

\paragraph{Limitations and Future Work.} Our experiments focus on a fixed clinical concept vocabulary, a single imaging modality and dataset, and one vision–language model for semantic evaluation. Future work will explore extending CARPA to broader concept spaces, additional modalities, datasets and alternative image editing pipelines. We will also conduct a more extensive expert evaluation.

\paragraph{Funding.} This work was supported by the Engineering and Physical Sciences Research Council (EPSRC) through the Causality in Healthcare AI Hub (CHAI), grant number EP/Y028856/1.

\bibliographystyle{named}
\bibliography{ijcai26}

\end{document}